# Evolving Neural Networks with Optimal Balance between Information Flow and Connections Cost

**Abdullah Khalili[1*], Abdelhamid Bouchachia[1]**

[1]Department of Computing and Informatics, Bournemouth University, United Kingdom.
Email: a.m.khalili@outlook.com.

*Corresponding author*

**Abstract** Evolving Neural Networks (NNs) has recently seen an increasing interest as an alternative path that might be more successful. It has many advantages compared to other approaches, such as learning the architecture of the NNs. However, the extremely large search space and the existence of many complex interacting parts still represent a major obstacle. Many criteria were recently investigated to help guide the algorithm and to cut down the large search space. Recently there has been growing research bringing insights from network science to improve the design of NNs. In this paper, we investigate evolving NNs architectures that have one of the most fundamental characteristics of real-world networks, namely the optimal balance between the connections cost and the information flow. The performance of different metrics that represent this balance is evaluated and the improvement in the accuracy of putting more selection pressure toward this balance is demonstrated on three datasets.

**Keywords** Neural Networks, Deep Learning, Neuroevolution, Complex Systems, Network Science, Graph Theory.

## 1. INTRODUCTION

Deep learning has recently revolutionized many fields. It has achieved remarkable advances and was able to address various challenging problems. The dominant approach is to design these architectures manually. Another interesting path seeks to evolve NNs automatically [1-3] this path is inspired by evolution and might be more successful. Recently, many researchers start to build neural network architectures automatically [4-8, 100]. The state of the art models on two popular benchmarks ImageNET and CIFAR were produced by search algorithms [6].

The main challenge to evolve NNs is the extremely large search space where there are dozens of complex interacting parts. Many researchers investigated using various criteria to help guide the search algorithm and to cut down the large search space. For examples, some works rewarded producing sparsity, modularity, regularity, and hierarchy [9-11]. Other works encouraged producing novelty, diversity, intrinsic motivation [12-14].

One challenge affecting the convergence of training NNs is the vanishing gradients problem [15-16]. To address this challenge, skip connections were introduced in [17] to maximize information flow. The model achieved the first place in the ILSVRC 2015 competition. To ensure maximum information flow of the network, Huang et al. [18] connected each layer to all subsequent layers and received inputs from all preceding layers. However, no optimal way to maximize information flow was proposed in these studies. Furthermore, adding this number of connections will largely increase the complexity of the network. Building networks that have optimal balance between connections cost and information flow has been largely investigated in graph theory and network science and there are many measures hypothesized to produce this balance.

Lately, there has been increasing use of tools from network science and graph theory because of their capability to characterize the behavior of the brain [19-23], these characteristics could help in building better NNs architectures and to avoid other characteristics related to different diseases and disorders. For example, many studies suggested an association between small-world characteristics and general intellectual abilities [24-26]. Langer et al. [25] showed that there is a significant association between the IQ and the centrality of nodes in the parietal cortex. Other studies suggested that the rich club coefficient and local efficiency continue to increase until adulthood in healthy subjects and then decrease with aging [27-28]. Petruo et al. [29] showed that the decline in performance caused by fatigue is associated with a decrease in global efficiency and small-world.

Metrics from network science and graph theory have also been used to diagnose different neurological and psychiatric disorders. Using the brain network characteristics, different studies [30-33] showed that patients with Alzheimer's Disease (AD) have decreased clustering coefficients and path lengths close to random networks. Rudie et al. [34] and Keown et al. [35] showed that patients with Autism Spectrum Disorder (ASD) have relatively lower modularity, clustering coefficient, and local efficiency (inefficiency of information flow in modules) while global efficiency is increased. Wang et al. [36] analyzed the rs-fMRI of patients with Attention-Deficit/Hyperactivity Disorder (ADHD), the results show that the brain networks of ADHD children is more-regular with increased local efficiency and a trend toward lower global efficiency in comparison with healthy subjects, indicating a delay in the developmental of the whole-brain functional networks.

Recently, there has been growing research bringing insights from graph theory and network science [37-39] to the design of NNs. Mocanu et al. [37] proposed a new approach called sparse evolutionary training (SET) to speed up the training process by enforcing small-world [40] and scale-free [41] properties. The method replaces the fully-connected layers with sparse layers before the training. They were able to quadratically reduce the number of parameters, with no decrease in performance. Mocanu et al [38] also showed that by constraining Restricted Boltzmann Machines (RBMs) and Gaussian RBMs (GRBMs) to a scale-free network, the number of parameters was reduced by a few orders of magnitude, with no loss in the accuracy. You et al. [75] studied the relationship between the predictive performance of a neural network and its graph structure. They found that neural network's predictive performance is approximately a smooth function of the clustering coefficient and the average path length of its relational graph. Many other recent studies [65-67] investigated to what extent network topology found in nature can lead to beneficial aspects in building NNs.

In this paper, we will investigate the use of measures from graph theory and network science that are evident in different biological and physical systems, such as local efficiency, global efficiency, eigenvector centrality, and entropy to characterize information flow, and the number of links or connections to characterize the connections cost. More specifically, we will investigate evolving NNs architectures that have an optimal balance between information flow and connections cost, several measures hypothesized to satisfy this balance will be evaluated.

The main contribution of this paper is showing that putting more selection pressure toward the balance between information flow and connections cost has improved the accuracy of the evolved NNs with relatively no increase in the architecture complexity. Different measures from network science were evaluated and compared after integrating them into the NEAT framework. We also analyze and compare the effect of different levels of selection pressure on the performance. Finally, we show how the performance of the evolved networks correlates with different measures when no selection pressure is used, i.e. when the performance alone is used as a fitness function.

The paper is organized as follows: Section 2 presents the evolutionary algorithm and the used fitness functions. Section 3 describes the experiment setup and the used datasets. Section 4 presents and analyzes the results. Section 5 discusses and concludes the paper with future research directions.

## 2. Method

Neuroevolution has many important features compared to gradient-based approaches, such as learning the architecture of the neural network [98, 99]. We will use the NEAT framework [42] described below to gradually increase the complexity of the neural network by introducing more structures (neurons and connections). Through this evolution process, we will test putting selection pressure toward the optimal balance between connections cost and information flow using different measures from graph theory and network science.

Keeping the connections cost low is related to the principle of simplicity in science [76-87], biological systems on the other hand seem to get more complex [88-90] because there are always rooms for more complex systems to account for all the complexity and challenges of the environment. Many studies in artificial systems also imply that a minimal level of complexity is necessary to attain a particular level of fitness [91-97].

Decreasing the number of links or connections in real-world networks has a clear advantage in reducing the cost whether in transport networks, human brains, etc. On the other hand, increasing information flow by introducing more links or connections has an important advantage in reducing the delay and the energy required to transfer information. In NNs, increasing information flow by introducing skip connections was found to be a better strategy than connecting each layer to the next layer only [17-18].

Although architectures with low connections cost are often desirable, putting selection pressures toward this criterion alone might result in architectures that are not efficient enough for information flow. On the other hand, increasing information flow alone might results in over-complex architectures. Therefore, we will investigate putting selection pressure toward the optimal balance between these two important criteria using different available measures. In the next section, we will give a description of the evolutionary algorithm and the used fitness functions.

**2.1 Evolutionary algorithm**

**A. Neuro Evolution of Augmenting Topologies (NEAT)**

The NEAT algorithm [42] will be used to evolve NNs architectures. The main reason behind using NEAT is its ability to start with a small network and optimize it, and then start adding new structures (neurons and connections) to the already optimized structure. This approach has many potential advantages, smaller structures can be optimized faster, where the algorithm can optimize the minimal structure needed to get a solution, then the increase in complexity resulting from introducing new structures will lead to new directions in the search space. The second benefit of increasing the complexity is that it provides a way out of local optima. Therefore, when a solution is on a local optimum, introducing new structures might lead to a way to escape this local optimum.

Mutation in NEAT can change both the connections weights and the network structure. Mutations in the structure can happen in two ways. In connection mutation, a new connection is added to connect two unconnected nodes. In node mutation, an existing connection is split into two connections and the new node is inserted between them. Only the structures that are found to be useful by fitness measures will be kept.

Small structures optimize faster than large structures, therefore adding more structure will initially decreases the fitness of the network. NEAT uses speciation to protect innovation and to avoid crossover between incompatible genomes, where the population is divided into species and networks with similar topologies are grouped in the same species. This will give the networks enough time to optimize their structures within their group and avoid competing with the entire population.

**B. Multi-objective optimization algorithm**

The evolutionary algorithm is based on a multiple objectives optimization algorithm [43, 10]. One advantage of using multi-objective algorithms instead of combining all objectives into a single function, is their ability to search all possible trade-offs between different objectives.

The algorithm uses Pareto dominance, where a solution $x^*$ dominate another solution $x$, if both the following conditions are satisfied: (1) $x^*$ is not worse than $x$ for any objective; (2) $x^*$ is better than $x$ for at least one objective. However, this treats all objectives equally, the performance should be more important than other objectives. To consider this, a stochastic version of Pareto dominance is used where other objectives are only considered with a probability $p$. Low values of $p$ will result in lower selection pressure on other objectives.

This stochastic application of other objectives is used as follow: $r$ is a random number between 0 and 1, and $p$ is the probability of considering other objectives. If $r > p$ then the candidate networks will be sorted according to the first objective only. If $r < p$ then the candidate networks will be sorted according to all objectives, i.e. if the candidate networks have the same first objective then sort them according to the second objective, and if they have the same second objective then sort them according to the third objective and so on. Putting large selection pressure toward other objectives might results in getting trapped in local optima, therefore a low $p$ value will be considered, where $p = 0.15$ is used.

**C. Mutation**

The used networks are feed-forward. Six types of mutation are possible, (1) 20% is the percentage of adding a single connection. (2) 20% is the percentage of removing a single connection. (3) 20% is the percentage of adding a single node. (4) 20% is the percentage of removing a single node. (5) 70% is the percentage that each node in the network to increment or decrement the bias, with both options equally probable. (6) 70% is the percentage that each connection in the network to increment or decrement the weigh, with both options equally probable. Table 1 summaries the parameters of the evolutionary algorithm.

**Table 1. Evolutionary algorithm parameters**

| Parameter | Value |
|---|---|
| Connection adding/deleting probability | 0.2 |
| Neurons adding/deleting probability | 0.2 |
| Weigh mutation probability | 0.7 |
| Bias mutation probability | 0.7 |

**2.2 The initial neural network model**

The initial neural network is a simple network that has 1 hidden layer with 12 hidden units. The main reason why we started with this simple network and not with an empty network is to speed up the evolution process by starting with a minimal network. Each input neuron is connected to all hidden neurons, and each hidden neuron is connected to all output neurons. The Sigmoid function is used as the activation function. Table 2 summarizes the parameters of the neural network.

**Table 2. Neural network parameters**

| Parameter | Value |
|---|---|
| Number of hidden neurons | 12 |
| Activation | Sigmoid |
| Initial connection | Each input neuron is connected to all hidden neurons, and each hidden neuron is connected to all output neurons |
| Network type | feed-forward |

**2.3 Fitness functions**

**A. Local and global efficiency**

Efficiency of a network measures how efficiently information flows in the network [50]. The Global Efficiency (GE) is the average inverse shortest path length in the network. The Local Efficiency (LE) measures information flow in sub-networks.

$$E(G) = \frac{1}{N(N-1)} \sum_{i \neq j} \frac{1}{d_{ij}} \qquad (4)$$

We will use two variants of this metric, the first one will divide the efficiency by the connections cost to achieve the balance between maximizing the efficiency and minimizing the connections cost. The second one will give more importance to the efficiency and will only take the connections cost in consideration when different networks have the same efficiency, i.e. when different models have the same efficiency, take the one with minimum connections cost.

**B. Centrality**

Centrality measures the importance or the influence of a node in a network; it can be also used to measure information flow in a network [51]. Degree Centrality (DC) is defined as the number of links incident upon a node [52]. Eigenvector Centrality (EC) of a node is calculated based on the centrality of its neighbours [53]. EC measures how well a node is connected to other well-connected nodes. EC of a node i, can be defined as the weighted sum of the centralities of all nodes connected to it by an edge $A_{ij}$.

$$EC_i = \frac{1}{\lambda} \sum_j A_{ij} V_j \quad (5)$$

Where V is the eigenvector associated to the eigenvalue $\lambda$ of A. We will use two variants of this metric, the first one will divide the centrality by the connections cost to achieve the balance between maximizing the centrality and minimizing the connections cost. The second one will give more importance to the centrality and will only take the connections cost in consideration when different models have the same centrality, i.e. when different networks have the same centrality, take the one with minimum connections cost.

**C. Statistical mechanics based measure**

Finally, we will use one of the most fundamental measure inspired from statistical mechanics [56-58], which seeks to maximize the entropy of the degree distribution of the network giving that the connections cost is constant. Maximizing the entropy of the degree distribution means that we encourage neurons to have more connections and hence improves information flow. The fixed connections cost constraint will result in a balance between regular lattice network (low entropy – low information flow – low cost) and random network (high entropy – high information flow – high cost). Therefore, among all the networks that have the same connections cost we will select the models with the higher information flow, i.e. higher entropy. The connections cost and Shannon entropy are given in (7) and (8) respectively.

$$\text{Connections Cost} = \sum_i n_i x_i \quad (7)$$

$$\text{Shannon Entropy} = -\sum_i p_i \log(p_i) \quad (8)$$

Where $n_i$ is the number of links at the $x_i$ degree, and $p_i = n_i/N$, where N is the total number of links.

**3. Experiments**

The following python libraries are used to evaluate the proposed measures: NEAT [59], powerlaw [60], and NetworkX [61]. Twenty runs were performed to evaluate each measure. We will test the proposed approach on the following three examples. The first one is the Cartpole example from OpenAI Gym gaming environment [62]. The aim is to prevent the pole from falling over. The gravity constant was changed to 26 to make the problem more difficult. Table 3 shows the parameters of the evolutionary algorithm. The second dataset is the iris flowers dataset [63]. The task is to predict the class of the flower. Table 4 shows the parameters of the evolutionary algorithm. The third dataset is the Wisconsin Breast Cancer Dataset [64] which contains two types of breast cancer. The task is to predict the class of the cancer. Table 5 shows the parameters of the evolutionary algorithm.

**Table 3. Evolutionary algorithm parameters for Cartpole example**

| Parameter | Value |
|---|---|
| Population size | 150 |
| Number of generations | 5000 |
| Simulation time | 60 |
| Gravity constant | 26 |

**Table 4. Evolutionary algorithm parameters for Iris flowers dataset**

| Parameter | Value |
|---|---|
| Population size | 10 |
| Number of generations | 1000 |

**Table 5. Evolutionary algorithm parameters for Wisconsin Breast Cancer Dataset**

| Parameter | Value |
|---|---|
| Population size | 10 |
| Number of generations | 1000 |

## 4. Results

**A. Accuracy and architecture complexity**

Firstly, we will show the performance of using the measures described in the earlier section and compare them to the results of using the performance alone as a fitness function. Table 6, Table 7, and Table 8 show the results for the three examples respectively. The measures that outperformed the baseline on all the three examples are highlighted, we can see that local and global efficiency, eigenvector centrality, and entropy show a clear improvement in the test accuracy for the three datasets with relatively close number of connections and neurons. However, no single measure has outperformed all other measures on all datasets, i.e. the results show that no particular way of increasing information flow is better than the others. We can also see that dividing the metrics by the connections cost is worse than using the connections cost as a separate objective, this might indicates that information flow should be given more importance than connections cost during the evolution process, where using the connections cost as a separate objective means that it will only be taken into account when other objectives are the same (+ means maximizing and − means minimizing in the below table, LE+Cost- for example means maximize LE as a first objective and minimize Cost as a second objective). The results show that the connections cost stays relatively close to the performance only case, which means that the evolved architectures have mainly changed the distribution of these connections to increase information flow.

**Table 6. The results of using different measures for the Cartpole example**

| Measure | Training Accuracy | Test Accuracy | Average Number of Connections | Average Number of Nodes |
|---|---|---|---|---|
| Performance Only | 85.26% | 45.68% | 24 | 12.5 |
| **LE+Cost-** | **80.47%** | **50.65%** | **24.2** | **11.9** |
| (LE/Cost)+ | 70.52% | 45.98% | 17.4 | 10.8 |
| **GE+Cost-** | **80.34%** | **51.04%** | **23.85** | **12.65** |
| (GE/Cost)+ | 80.34% | 45.67% | 21.6 | 12.05 |
| DC+Cost- | 65.67% | 41.80% | 23 | 11.45 |
| (DC/Cost)+ | 85.28% | 51.44% | 23.95 | 14 |
| **EC+Cost-** | **90.17%** | **60.27%** | **27.3** | **12.95** |
| **(EC/Cost)+** | **90.19%** | **56.99%** | **34.4** | **16.25** |
| **Entropy+Cost-** | **75.42%** | **46.95%** | **24** | **12** |
| (Entropy/Cost)+ | 85.28% | 56.26% | 25.45 | 12.65 |

**Table 7. The results of using different measures for the Iris flowers dataset**

| Measure | Training Accuracy | Test Accuracy | Average Number of Connections | Average Number of Nodes |
|---|---|---|---|---|
| Performance Only | 98.67% | 89.33% | 80.65 | 19.3 |
| **LE+Cost-** | **99.13%** | **91.13%** | **74.35** | **18.95** |
| (LE/Cost)+ | 98.85% | 88.98% | 77.05 | 18.90 |
| **GE+Cost-** | **98.98%** | **89.60%** | **79.25** | **19.5** |
| (GE/Cost)+ | 98.93% | 87.32% | 69.50 | 17.95 |
| DC+Cost- | 98.99% | 89.12% | 73.65 | 18.75 |
| (DC/Cost)+ | 98.71% | 89.28% | 77.35 | 19.2 |
| **EC+Cost-** | **98.85%** | **91.17%** | **74.15** | **19.25** |
| **(EC/Cost)+** | **98.79%** | **89.73%** | **79.85** | **19.25** |

| Entropy+Cost- | 99.01% | **90.09%** | 71.05 | 18.35 |
| (Entropy/Cost)+ | 98.89% | 89.12% | 74.6 | 19.55 |

**Table 8.** The results of using different measures for the breast cancer dataset

| Measure | Training Accuracy | Test Accuracy | Average Number of Connections | Average Number of Nodes |
|---|---|---|---|---|
| Performance Only | 98.77% | 92.68% | 120.4 | 23 |
| **LE+Cost-** | **98.68%** | **93.76%** | **121.5** | **23.1** |
| (LE/Cost)+ | 98.79% | 91.63% | 111.6 | 22.2 |
| **GE+Cost-** | **99.02%** | **93.73%** | **118.65** | **22.25** |
| (GE/Cost)+ | 98.9% | 94.67% | 119.7 | 22.65 |
| DC+Cost- | 98.59% | 91.98% | 116.95 | 22.85 |
| (DC/Cost)+ | 98.65% | 96.98% | 123.75 | 23.65 |
| **EC+Cost-** | **98.9%** | **93.26%** | **126.25** | **23.5** |
| **(EC/Cost)+** | **98.93%** | **95.51%** | **118.5** | **22.85** |
| **Entropy+Cost-** | **98.95%** | **93.71%** | **126.7** | **22.9** |
| (Entropy/Cost)+ | 98.77% | 95.88% | 115.75 | 22.7 |

**B. The correlation between the accuracy and different measures**

Here we will study how different measures change during the evolution process. Figure 2 to Figure 6 show how local and global efficiency, eigenvector centrality, entropy, and the number of connections change during the evolution process when the accuracy is used as the only fitness function. We can see a correlation between the increase in accuracy and eigenvector centrality, entropy, and the number of connections. However, this is not the case for global efficiency, this can be explained by the fact that the initial architecture has a maximum global efficiency, therefore we see a decrease in this measure. Although we started with a simple architecture, we can see that local efficiency is the only measure that shows strong fluctuations when the accuracy increases. This is not the case for other measures which almost show a continues increase or decrease. Due to this correlation, putting selection pressure toward these measures might help in finding a better trajectory in the search space.

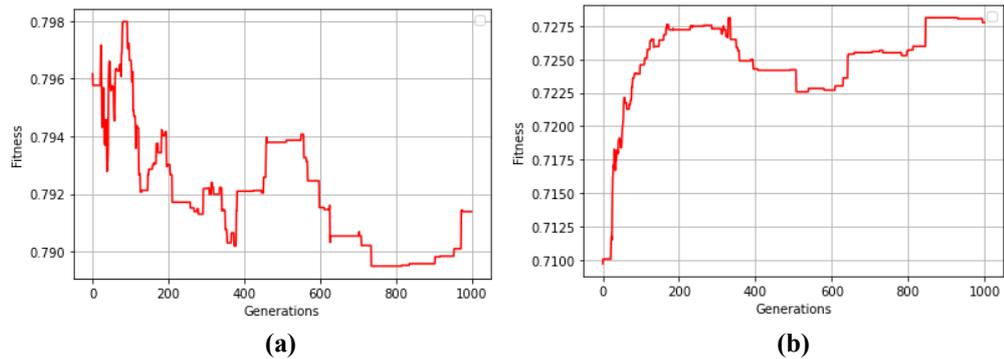

**(a)** **(b)**
**Figure 2.** The correlation between the performance and the local efficiency for (a) Iris Flowers, (b) Breast Cancer.

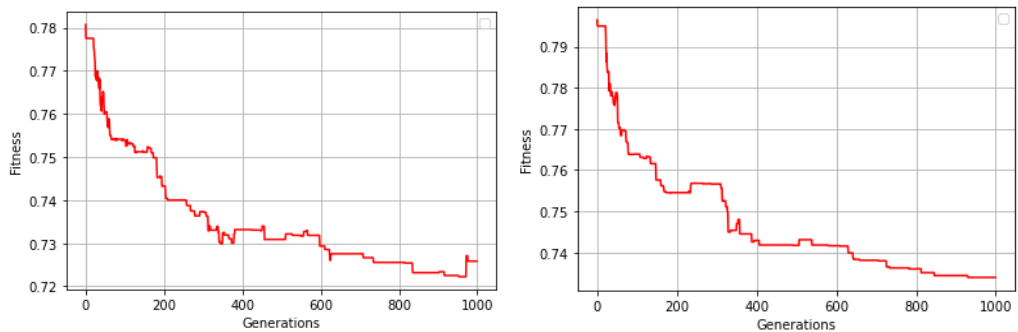

**(a)** **(b)**

**Figure 3.** The correlation between the performance and the global efficiency for (a) Iris Flowers, (b) Breast Cancer.

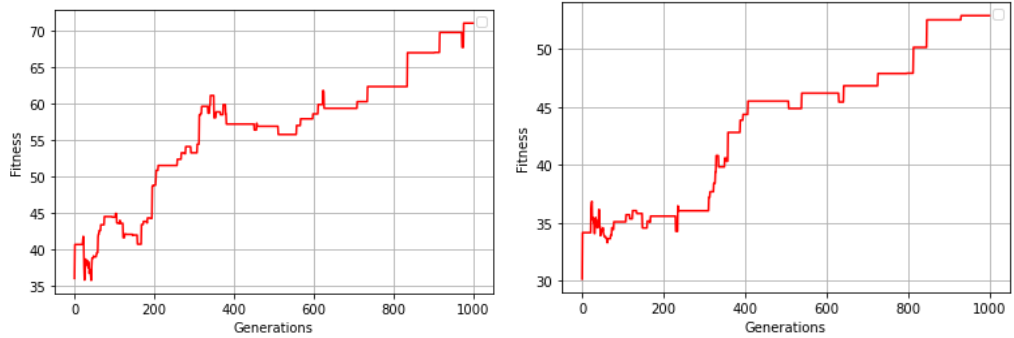

**(a)** **(b)**

**Figure 4.** The correlation between the performance and the eigenvector centrality for (a) Iris Flowers, (b) Breast Cancer.

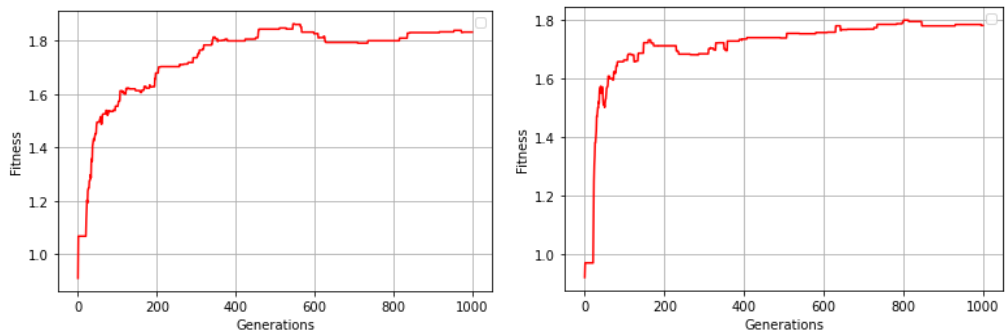

**(a)** **(b)**

**Figure 5.** The correlation between the performance and the entropy for (a) Iris Flowers, (b) Breast Cancer.

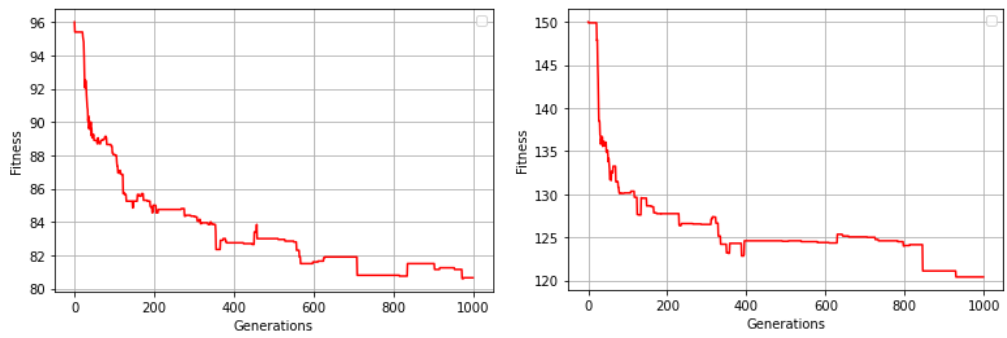

**(a)** **(b)**

**Figure 6.** The correlation between the performance and the number of connections for (a) Iris Flowers, (b) Breast Cancer.

**D. Sensitivity to higher selection pressure**

Here we will compare the performance for different levels of selection pressure when using the eigenvector centrality as a fitness function. Table 9 to Table 11 show that using a higher selection pressure has resulted in a lower test accuracy for the three examples, keeping the selection pressure low seems to be a better strategy to keep the search space more diverse and to avoid getting trapped in local optima. Due to the fact that only low selection pressure is used, this means that the final evolved architecture will not be necessarily optimal, but the selection pressure toward this optimal balance has helped in improving the accuracy, i.e. the selection pressure toward this balance has helped in finding a better trajectory in the search space.

**Table 9. The sensitivity to higher selection pressure for the Cartpole example**

| Measure | Training Accuracy | Test Accuracy | Average Number of Connections | Average Number of Nodes |
|---|---|---|---|---|
| P = 0 | 85.26% | 45.68% | 24 | 12.5 |
| **P = 0.15** | **90.17%** | **60.27%** | **27.3** | **12.95** |
| P = 0.50 | 65.59% | 37.73% | 20.5 | 11.3 |

**Table 10. The sensitivity to higher selection pressure for the Iris flowers dataset**

| Measure | Training Accuracy | Test Accuracy | Average Number of Connections | Average Number of Nodes |
|---|---|---|---|---|
| P = 0 | 98.67% | 89.33% | 80.65 | 19.3 |
| **P = 0.15** | **98.85%** | **91.17%** | **74.15** | **19.25** |
| P = 0.50 | 99.06% | 88.76% | 75.65 | 19.6 |

**Table 11. The sensitivity to higher selection pressure for the breast cancer dataset**

| Measure | Training Accuracy | Test Accuracy | Average Number of Connections | Average Number of Nodes |
|---|---|---|---|---|
| P = 0 | 98.77% | 92.68% | 120.4 | 23 |
| **P = 0.15** | **98.9%** | **93.26%** | **126.25** | **23.5** |
| P = 0.50 | 98.98% | 91.70% | 115.6 | 24.7 |

## 5. Discussion and Conclusion

Many recent studies have shown how different measures from network science and graph theory correlate to different characteristics and disorders of the human brain. Many of these measures have been also observed in different biological and physical networks. There are growing interest recently in investigating to what extent network topology found in nature can lead to beneficial aspects in building ANN.

In this paper, we investigated one of the most fundamental characteristics of real-world network, namely the balance between connections cost and information flow. Decreasing the connections cost is an important criterion in building ANN [68-74]. Increasing information flow is also an effective strategy in many successful neural network architectures. In this paper, we investigated the use of many measures hypothesised to produce the optimal balance between information flow and connections cost. The results were consistent for different examples and different measures, and showed a clear improvement in the accuracy particularly using local and global efficiency, eigenvector centrality, and entropy. This shows that putting selection pressure toward this balance has helped in guiding the algorithm and in cutting down the large search space. However, keeping the selection pressure low was essential to keep the population more diverse and to avoid local optima. The results also showed that there is a correlation between the increase in accuracy and eigenvector centrality, entropy, and the number of connections even when no selection pressure toward them is used.

This study along with many other recent studies have shown that using measures from graph theory and network science could have many advantages in building ANN. It represents a promising research direction that worth further investigations. Future work will investigate how well these measures scale with larger models and what is the best mechanism to integrate them in training deep neural networks. Future work will also investigate using other measures and optimization processes from network science and graph theory to further help in finding better architectures.